\documentclass{issw}

\usepackage[base]{babel}
\usepackage{graphicx}
\makeatletter
\@ifundefined{KV@Gin@alt}{\define@key{Gin}{alt}{}}{}
\makeatother
\hypersetup{pdflang={en}}
\usepackage{amsmath}

\usepackage{parskip}
\usepackage[acronym,nomain]{glossaries}

\newacronym{sar}{SAR}{synthetic aperture radar}
\newacronym{saraai}{SAR-AAI}{SAR-detected Avalanche Activity Index}
\newacronym{nwp}{NWP}{numerical weather prediction}
\newacronym{dem}{DEM}{digital elevation model}
\newacronym{nve}{NVE}{Norwegian Water Resources and Energy Directorate}
\newacronym{mlp}{MLP}{multilayer perceptron}
\newacronym{fc}{FC}{fully connected}
\newacronym{gelu}{GELU}{Gaussian error linear unit}
\newacronym{esa}{ESA}{European Space Agency}
\glsdisablehyper

\title{DATA-DRIVEN PREDICTION OF SATELLITE-OBSERVED AVALANCHE ACTIVITY FROM
SNOWPACK SIMULATIONS}

\author[1]{Jakob Grahn\corref{corresponding_author}}
\author[1,2]{Filippo Maria Bianchi}
\author[3]{Bert Kruyt}
\author[3]{Karsten M\"uller}

\address[1]{NORCE Research, Troms\o, Norway}
\address[2]{UiT The Arctic University of Norway, Troms\o, Norway}
\address[3]{Norwegian Water Resources and Energy Directorate, Oslo, Norway}

\cortext[corresponding_author]{
Jakob Grahn, NORCE,\\
9294 Troms\o, Norway; jgra@norceresearch.no}

\begin{document}
\pagestyle{empty}

\begin{abstract}
Avalanche forecasting requires knowledge of the snowpack and recent avalanche activity, but both are difficult to keep track of across large mountain regions.
Field observations are essential but often sparse, which limits regional monitoring and development of numerical or statistical prediction models.
Synthetic aperture radar (SAR) can repeatedly map avalanche debris over large areas, opening up new opportunities for data-driven approaches for avalanche forecasting.
In this study, we take a first step in this direction.
We constructed two large datasets for five winters with two operational
Sentinel-1 satellites and a typical six-day repeat interval. First, we mapped
avalanche debris in Sentinel-1 images across Norway and parts of Sweden.
Secondly, we ran the SNOWPACK model forced by numerical weather predictions on a 20$\times{}$20 km grid, at different elevations and predefined slope angles.
A transformer was then trained to use five days of SNOWPACK outputs to predict activity mapped by SAR for the following day.
We represented activity with the SAR-detected Avalanche Activity Index (SAR-AAI), a study-specific index that gives larger debris more weight, spreads detections across possible occurrence dates and normalises by modelled runout area.
We trained the transformer on four winters and evaluated it on one. 
Regional mean predicted and reference SAR-AAI had a Pearson correlation of 0.803 when averaged over complete six-day periods, with each value placed at the midpoint of its period.
The model followed broad changes in time and space, but produced smoother predictions and underestimated the strongest activity. 
At the 20 km cell scale, agreement after the same six-day averaging was weaker ($r=0.549$).
The result is based on a single training run of the machine-learning model and has not been tested on an untouched winter. 
The SAR dataset is incomplete and contains detection errors and uncertain timing. 
Thus, the results do not yet show operational forecast skill.
Still, predictions based on regional SNOWPACK simulations followed broad changes mapped by Sentinel-1. 
This is a promising first step towards using SAR avalanche detections with snowpack modelling for avalanche forecasting.
\end{abstract}


\maketitle
\raggedbottom
\glsresetall

\section{INTRODUCTION}

Avalanche forecasting requires knowledge of the snowpack and recent avalanche
activity. Both are difficult to keep track of across large mountain regions.
Field observations are essential, but they are sparse, uneven and often
limited by visibility and access. This is particularly challenging in the
Scandinavian Mountains, where forecasters assess large and sparsely populated
areas.

Spaceborne \gls{sar} has become a complementary way to map avalanche activity.
Changes in radar backscatter between repeat-pass images can reveal avalanche
debris. Methods have progressed from manual inspection and simple thresholding
to automatic machine-learning methods that outline likely debris
\parencite{Vickers2016,Eckerstorfer2019,Bianchi2021}. \Gls{sar} does not give
a complete record of activity: avalanches can be missed, false detections can
remain, and the release time is only known to fall between two observations.
Even so, it provides repeated observations across areas that are difficult to
cover from the ground.

Snowpack modelling has developed in parallel. Physically based models such as
SNOWPACK use meteorological observations or numerical weather prediction to
simulate how the snowpack develops through the winter
\parencite{Bartelt2002,Herla2024}. \textcite{Herla2024} describe a regional
model chain that runs SNOWPACK across Norway at several elevations and
predefined slope aspects. These simulations provide consistent snowpack
histories where direct measurements are scarce.

\begin{figure*}[t]
  \centering
  \includegraphics[
    width=0.98\textwidth,
    alt={Timeline of operational Sentinel-1A, B, C and D data with a thin strip for December-to-April study winters. Blue winters from 2016--2017 through 2020--2021 had two satellites and were included. Grey single-satellite winters were excluded because the repeat interval increased from about six to twelve days. The orange hatched 2025--2026 winter had two satellites but no SNOWPACK simulation.}
  ]{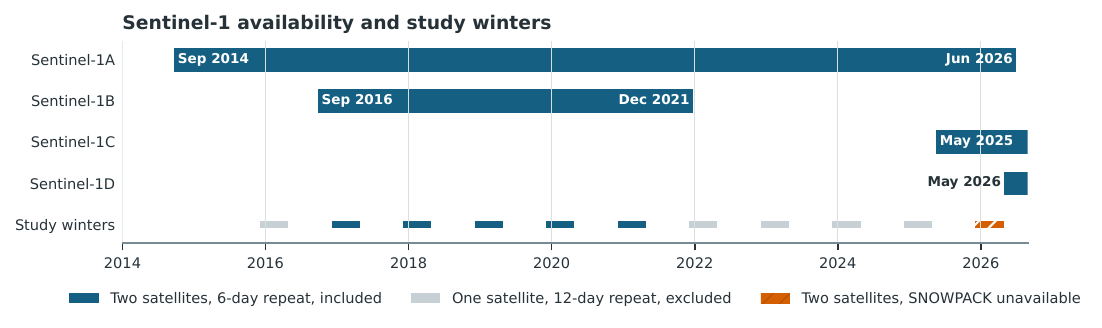}
  \caption{Sentinel-1 availability and study-season selection. The satellite
  lanes show operational data periods, rounded to the month and through August
  2026; the Sentinel-1A endpoint follows its announced phase-out. The thin
  study-winter strip marks each 1~December--30~April period. Blue winters had
  two satellites, a typical six-day repeat interval and matching SNOWPACK
  simulations, and were included. Grey winters had one satellite and a typical
  twelve-day interval. The orange hatched winter had two satellites but no
  SNOWPACK simulation. Mission availability does not guarantee an acquisition
  in every grid cell. Dates follow Copernicus mission records
  \parencite{copernicus2026sentinel1mission}.}
  \label{fig:sentinel-timeline}
\end{figure*}

Using historical data to predict avalanche conditions is not new. Early
statistical models used discriminant analysis and nearest-neighbour methods,
followed by classification trees and support-vector machines
\parencite{Buser1983,Kronholm2006,Pozdnoukhov2011}. More recent studies have
used random forests and other machine-learning methods to predict regional
activity classes, natural dry- and wet-snow avalanche days, and danger levels
assigned by forecasting services
\parencite{Harvey2016,Sielenou2021,PerezGuillen2022,Viallon2023,
Mayer2023,Hendrick2023,Eiselt2025}. These studies show that meteorological and
simulated snowpack data contain useful information about avalanche conditions.
Most, however, use field observations, individual avalanche-path records or
forecaster assessments as their training target.

Here, we bring together two large regional datasets. The first is a
multiwinter catalogue of avalanche-debris detections from Sentinel-1 images
across Norway and parts of Sweden. The second contains NWP-driven SNOWPACK
simulations on a 20 by 20~km grid, resolved by elevation and slope aspect.
Together, they allow us to test whether recent simulated snowpack conditions
can predict the following day's activity mapped by \gls{sar}.

\textcite{grahn2024forecasting} described an earlier version of our
\gls{sar} dataset and outlined a different, pixel-scale model based directly
on \gls{nwp} fields and previous avalanche masks. Here, we instead calculate
\gls{saraai} for each 20~km cell and predict it from five days of regional
SNOWPACK histories. To our knowledge, this is the first study to use a
multiwinter, wide-area \gls{sar}-derived avalanche-activity dataset as the
prediction target for regional SNOWPACK simulations resolved by elevation and
slope aspect. This is a first experiment rather than an operational
forecasting system.

\section{DATA}

\subsection{Satellite-observed avalanche activity}

We produced the avalanche dataset from repeat-pass Sentinel-1 \gls{sar}
images. A segmentation model compared radar backscatter from consecutive
passes, together with terrain information, and outlined likely avalanche debris
\parencite{bianchi2021segmentation}. Each image pair produced polygons of likely
debris. The same debris could appear in overlapping satellite views, so we
merged polygons that matched in space and time. A merged polygon represents
mapped debris, not necessarily a distinct avalanche. We defined its
\emph{occurrence interval}, the period when the avalanche could have occurred,
as the intersection between the possible occurrence periods of the merged
polygons. We did not assign an exact release date.

As Figure~\ref{fig:sentinel-timeline} shows, we used five complete
two-satellite winters, 2016--2017 through 2020--2021. One Sentinel-1 satellite repeats a given
orbit after 12 days. With two satellites offset in the same orbit, the interval
is 6 days. The shorter interval improves the timing of satellite-mapped
activity, as well as detection capabilities. We ran the
detection algorithm from 1~December to 30~April each winter. Excluding late
spring reduced likely bare-ground and
wet-snow false positives. Even within this period, no detection can mean either
that no avalanche was mapped or that no suitable observation was available.
\Gls{sar} mapping can also miss small, dry or loose avalanches. We therefore aknowledge that the
dataset is an incomplete record of avalanche activity, and should be used as an indicator.

We screened the merged detections by polygon size, elevation and land cover. 
The avalanche-debris polygons were at least 2,000~m$^2$ and smaller than
200,000~m$^2$ and had a mean \gls{dem} elevation above 75~m. A polygon was
rejected when at least 25\% of its area was classified as cropland, built-up
land or coniferous forest, or when at least 50\% was permanent water. These
rules were applied only where the land-cover product completely covered and
classified the polygon. During daily target construction, each assigned date
also required at least 0.30~m of simulated snow at the detection's elevation
and aspect. This removes likely snow-free false positives, but uses one model
input to screen the reference data.

\subsection{Regional SNOWPACK simulations}

We ran SNOWPACK \parencite{bartelt2002snowpack} with reanalysis from MET
Norway's Long-Term Consistent (LTC) archive on the 1~km MET Nordic grid. AWSOME used the median values within each
20 by 20~km cell and 300~m elevation band \parencite{herla2024modelchain}.
For the five study winters, we used 2,589 available cell and elevation
combinations. Each cell used up to seven elevation levels from 300 to 2,100~m.
At each level, one flat simulation and four simulated slopes at
38$^{\circ}$ facing north, east, south and west represented solar and wind
exposure.

The configuration was adapted with the \gls{nve} for Norwegian conditions.
Changes covered how new-snow density and grain shape were calculated, along
with erosion, redeposition and snow exchange between the simulated slopes.

The study period ends in 2020--2021. No matching 2025--2026 SNOWPACK simulation
was available at the time of writing, although that winter met the satellite
requirement (Figure~\ref{fig:sentinel-timeline}).

\subsection{Terrain data}

We used a \gls{dem} and a modelled flow-depth map to identify avalanche runout
terrain within each grid cell. We modelled runout with MoT-Voellmy
\parencite{issler2025motvoellmy}. The model used terrain downsampled to 20~m, and
we returned its output to the 10~m \gls{dem} grid.
We counted pixels with maximum flow depth above 0.1~m as runout terrain and
summed them to obtain the total modelled runout area in each cell. We
marked targets invalid in cells with less than 10,000~m$^2$ of
modelled runout terrain. Runout information only defined the target and which
cells were valid; it was not a model input. Dividing by runout area expresses mapped
activity per unit of modelled runout terrain, and the minimum prevents ratios
based on very small areas.

\section{METHODS}

We assigned detections to possible avalanche dates, calculated daily
\gls{saraai}, as defined in equation~\eqref{eq:saraa}, and trained a transformer.

\subsection{Input features: five-day SNOWPACK history}

Each sample represented one 20~km grid cell and target date. The input
comprised the five preceding daily SNOWPACK states for up to five slope
directions and seven elevation levels.

Archived SNOWPACK profiles were normally available every three hours, with
occasional gaps. We reduced each variable-length layer profile to 277 fixed
features: layer count, profile height and total thickness; the mean, standard
deviation, minimum and maximum of numerical layer properties; surface and basal
layer values; and grain-type fractions and indicators.
We kept the latest profile for each local day and a
mask identified missing day--slope--elevation combinations. The model therefore
received one summary vector per daily profile, not the ordered snow
layers or changes within the day.

\subsection{Target labels: following-day SAR-AAI}

We overlaid each avalanche-debris polygon on the 20~km grid. To screen likely
snow-free detections, we matched each polygon to the SNOWPACK simulations for
its mean elevation and terrain orientation. We kept only possible dates with at
least 0.30~m of simulated snow. Because the release date was unknown, we spread
each detection equally across the possible dates that passed the seasonal and
snow-depth filters. We then summed all contributions into one value for each
cell and date. The daily index therefore spreads the mapped activity across the
dates when the avalanche may have occurred; it does not identify the release
date.

The target for each sample was one \gls{saraai} value for the whole cell on the
following day. This study-specific index is defined precisely for cell $c$ and
date $d$ in Equation~\eqref{eq:saraa}:

\begin{equation}
\mathrm{SAR\text{-}AAI}_{c,d}=
\frac{10^{6}}{R_c}
\sum_{i\in(c,d)}
\left(\frac{A_i}{10^{4}}\right)^{1.5}
\frac{f_{i,c}}{n_i},
\label{eq:saraa}
\end{equation}

Equation~\eqref{eq:saraa} sums all detections that overlap cell $c$ and could
have occurred on date $d$. For each detection, $A_i/10^4$ expresses its full
mapped debris area relative to 10,000~m$^2$, equivalent to a
$100~\mathrm{m}\times100~\mathrm{m}$ reference area. The exponent 1.5 gives
larger polygons more weight. We chose it during exploratory work because larger
avalanches are generally more dangerous and therefore more important to
predict, but we did not calibrate it from physical data. The factor
$f_{i,c}$ is the fractional overlap of the detection with cell $c$. 
The divisor $n_i$ spreads it equally across the possible dates that passed the seasonal and snow-depth
filters: $n_i=1$ assigns the full weight to one date, $n_i=2$ assigns half to
each of two dates, and so on. Finally, $10^6/R_c$ expresses the summed weight
per square kilometre of modelled runout terrain in the cell, the index represents 
activity in units of actual avalanche terrain. 

\Gls{saraai} is a study-specific index, not an avalanche count, conventional
avalanche-size scale or operational danger level. It describes mapped debris
rather than release activity, and its timing remains uncertain. We refer to the
label as reference \gls{saraai} and the model output as predicted \gls{saraai}.

We included a cell-winter only if it contained at least one positive detection.
This excluded quiet or unobserved cases, which an operational system would
also need to handle.

\subsection{Transformer architecture and training}

Each daily 277-feature SNOWPACK summary was linearly encoded, then combined
with learned representations of its position within the five-day history,
slope aspect and elevation. Two transformer layers
\parencite{vaswani2017attention} processed the available encoded states, which
attention pooling combined into one representation for the cell. An
availability mask prevented missing day--aspect--elevation combinations from
contributing to the transformer or pooling. Input features were scaled using
the training seasons, and missing values were set to zero.

The model had two output heads. A classification head predicted whether
\gls{saraai} was present, while a regression head predicted its positive value.
The model received no explicit cell identifier, coordinates, runout area or
historical activity statistic for the cell.

Weighted binary cross-entropy trained the classification head. The regression
head was trained on positive reference \gls{saraai} using Smooth-L1 loss on the
transformed target and symmetric mean-squared error in the original
\gls{saraai} scale. We used AdamW with dropout, learning-rate reduction and
early stopping. Early stopping and model selection used validation RMSE for
positive \gls{saraai} after returning predictions to the original scale.
Classification probabilities below 0.5 gave zero predicted \gls{saraai}.

We trained on 332,623 samples from 2016--2017, 2017--2018, 2018--2019 and
2020--2021. We validated on 102,448 samples from 674 cells in 2019--2020. The
split was seasonal, not spatial, and 642 validation cells also appeared in the
training winters.

\begin{figure*}[p]
  \centering
  \includegraphics[
    width=0.95\textwidth,
    alt={Regional time series and a density plot for individual cells and dates in the 2019--2020 validation season. The predicted SAR-AAI follows most broad reference SAR-AAI cycles but underestimates the largest peak. High-reference points mainly fall below the one-to-one line.}
  ]{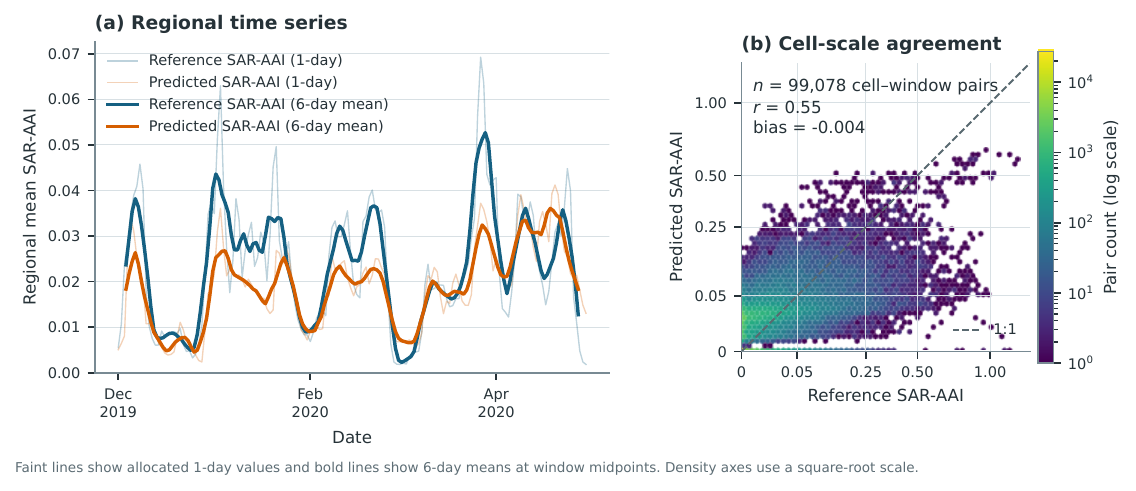}
  \caption{Regional and cell-scale \gls{saraai} in the 2019--2020 validation
  season. (a) Mean reference and predicted \gls{saraai} across 674 cells;
  faint lines are daily values and bold lines are six-day means placed at the
  midpoint of each period. (b) Density of 99,078 cell and window combinations
  from 147 periods with all six days available. The axes use a square-root scale labelled in
  \gls{saraai} units; colour is logarithmic.}
  \label{fig:temporal-density}
\end{figure*}

\begin{figure*}[p]
  \centering
  \includegraphics[
    width=0.95\textwidth,
    alt={Six maps compare reference SAR-AAI and predicted SAR-AAI for low, typical and peak six-day periods. Broad spatial concentrations are similar, while predicted SAR-AAI is smoother and weaker during the peak period.}
  ]{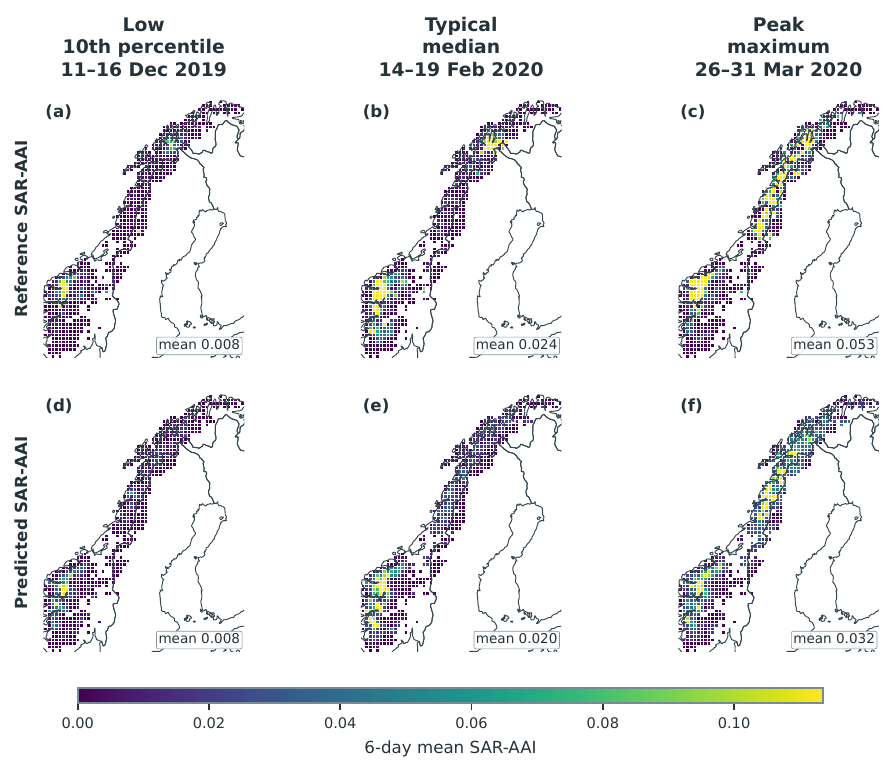}
  \caption{Six-day mean \gls{saraai} for the periods nearest the 10th
  percentile (low), median (typical) and maximum (peak) regional reference
  value. Reference and predicted values share one colour range; grey cells
  have no validation sample. Borders: Natural Earth Admin~0 Countries,
  1:50m, version 5.1.1.}
  \label{fig:spatial-comparison}
\end{figure*}

\section{RESULTS}

Because \gls{sar} timing is uncertain, we compare the regional series and
values for individual cells using complete six-day averages, matching the
maximum occurrence interval. We place each average at the midpoint of its
period, halfway between the two middle dates (Figure~\ref{fig:temporal-density}).
This smoothing is used only for evaluation and plotting; it does not change the
daily model output. Across 147 validation periods, regional mean predicted
\gls{saraai} followed the main reference \gls{saraai} cycles, with Pearson
correlation 0.803. During the peak period, 26--31~March 2020, predicted
\gls{saraai} reached 60.0\% of the reference.

At cell scale, the predictions showed more scatter and followed reference
\gls{saraai} less closely. Across 99,078 cell and window combinations with
complete six-day averages, Pearson correlation was 0.549. Most high reference
values lay below the one-to-one line, showing that the model underestimated the
strongest activity.

Across the representative low, typical and peak periods mapped in
Figure~\ref{fig:spatial-comparison}, the predictions reproduced several broad
concentrations of activity. They were smoother than the reference and
slighlty weaker during the March peak. The model however reproduced large spatial
patterns within the season, but it missed some local concentrations.

\begin{figure}[!t]
  \centering
  \includegraphics[
    width=\columnwidth,
    alt={Four aligned time series compare all validation cells with low-, medium- and high-activity regimes defined from training data. The shared square-root vertical scale expands low values while retaining one scale across panels. Timing agrees most clearly for medium- and high-activity cells.}
  ]{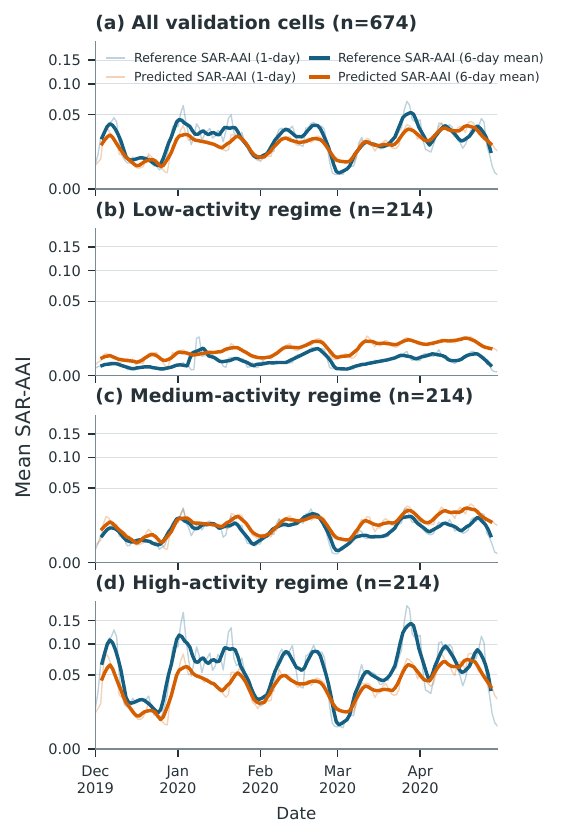}
  \caption{Daily and six-day mean reference and predicted \gls{saraai} for all
  674 validation cells and three equal activity regimes based on average reference
  activity in the training winters (214 cells each). Validation values did not
  define the regimes. Panels share a square-root scale.}
  \label{fig:stratified-comparison}
\end{figure}

We divided the 642 cells also present in the training seasons into equal-sized
low-, medium- and high-activity regimes using only their mean reference
\gls{saraai} in the training seasons
(Figure~\ref{fig:stratified-comparison}). For the low-, medium- and high-activity
regimes, correlations between the six-day reference and predicted series were
0.563, 0.766 and 0.813, respectively. The predicted \gls{saraai} followed the
timing most clearly in the medium- and high-activity regimes. It underestimated
high-activity peaks and tended to exceed the low-activity reference late in the
season.

\section{DISCUSSION AND CONCLUSIONS}

We tested whether five days of regional, terrain-resolved SNOWPACK simulations
could predict the following day's avalanche activity, represented by the
\gls{saraai}. When averaged over six days and across the region, the
predictions followed the main activity cycles in the \gls{sar} dataset. During
the 2019--2020 validation season, regional predicted and reference
\gls{saraai} had a correlation of 0.803. The model reproduced broad changes in
time and space, but produced smoother maps, underestimated the strongest
activity and reached about 60\% of the reference during the peak period.
Agreement was weaker at the 20~km cell scale, suggesting that the model
captured regional patterns better than local extremes.

This is an encouraging result because the model used only recent SNOWPACK
states resolved by elevation and slope aspect. It received no explicit
coordinates, runout area or history of avalanche activity. The agreement
therefore indicates that the regional snowpack simulations and the separate
satellite dataset contain some of the same broad changes in avalanche
conditions.

Part of this relationship may reflect conditions that affect whether avalanche
debris is visible in repeat-pass \gls{sar}, as well as conditions associated
with avalanche release. The results should therefore be interpreted as a
prediction of satellite-observed activity. Even so, the results indicate a
promising connection between wide-area \gls{sar} detections and physically
based snowpack simulations. This provides a useful basis for developing
regional models of avalanche activity and, eventually, for supporting
avalanche forecasting.

\phantomsection
\addcontentsline{toc}{section}{ACKNOWLEDGEMENTS}
\section*{ACKNOWLEDGEMENTS}

This work was supported by the \gls{esa} through PRODEX contract 4000137043
(AFEX), administered by the Norwegian Space Agency. We thank MET Norway for
meteorological data and the Norwegian Mapping Authority for elevation data.

\printbibliography[title={REFERENCES}]

@article{bianchi2021segmentation,
  author       = {Bianchi, Filippo Maria and Grahn, Jakob and Eckerstorfer,
                  Markus and Malnes, Eirik and Vickers, Hannah},
  title        = {Snow Avalanche Segmentation in {SAR} Images With Fully
                  Convolutional Neural Networks},
  journaltitle = {IEEE Journal of Selected Topics in Applied Earth Observations
                  and Remote Sensing},
  year         = {2021},
  volume       = {14},
  pages        = {75--82},
  doi          = {10.1109/JSTARS.2020.3036914}
}

@article{bartelt2002snowpack,
  author       = {Bartelt, Perry and Lehning, Michael},
  title        = {A physical {SNOWPACK} model for the Swiss avalanche warning:
                  Part {I}: Numerical model},
  journaltitle = {Cold Regions Science and Technology},
  year         = {2002},
  volume       = {35},
  number       = {3},
  pages        = {123--145},
  doi          = {10.1016/S0165-232X(02)00074-5}
}

@inproceedings{grahn2024forecasting,
  author    = {Grahn, Jakob and Bianchi, Filippo Maria and M{\"u}ller, Karsten
               and Malnes, Eirik},
  title     = {Data-driven avalanche forecasting -- using weather and satellite
               data},
  booktitle = {Proceedings of the International Snow Science Workshop},
  year      = {2024},
  address   = {Troms{\o}, Norway},
  pages     = {39--44}
}

@inproceedings{herla2024modelchain,
  author    = {Herla, Florian and Widforss, Aron and Binder, Michael and
               M{\"u}ller, Karsten and Horton, Simon and Reisecker, Michael and
               Mitterer, Christoph},
  title     = {Establishing an operational weather \& snowpack model chain in
               Norway to support avalanche forecasting},
  booktitle = {Proceedings of the International Snow Science Workshop},
  year      = {2024},
  address   = {Troms{\o}, Norway},
  pages     = {168--175}
}

@online{copernicus2026sentinel1mission,
  author  = {{European Space Agency and European Commission}},
  title   = {Sentinel-1 Mission},
  year    = {2026},
  url     = {https://sentiwiki.copernicus.eu/web/s1-mission},
  urldate = {2026-08-28}
}

@report{issler2025motvoellmy,
  author      = {Issler, Dieter},
  title       = {Basic Equations and Numerical Methods in {MoT-Voellmy}},
  institution = {Norwegian Geotechnical Institute},
  type        = {Technical Note},
  number      = {20230100-06-TN},
  date        = {2025-02-20},
  url         = {https://github.com/norwegian-geotechnical-institute/MoT-Voellmy/blob/main/Documentation/20230100-06-TN.MoT-Voellmy_eqs_numerics.pdf},
  urldate     = {2026-08-31}
}

@inproceedings{vaswani2017attention,
  author    = {Vaswani, Ashish and Shazeer, Noam and Parmar, Niki and
               Uszkoreit, Jakob and Jones, Llion and Gomez, Aidan N. and
               Kaiser, {\L}ukasz and Polosukhin, Illia},
  title     = {Attention Is All You Need},
  booktitle = {Advances in Neural Information Processing Systems},
  year      = {2017},
  volume    = {30},
  pages     = {5998--6008},
  url       = {https://proceedings.neurips.cc/paper/2017/hash/3f5ee243547dee91fbd053c1c4a845aa-Abstract.html}
}

@article{Vickers2016,
  author       = {Vickers, Hannah and Eckerstorfer, Markus and Malnes, Eirik and Larsen, Yngvar and Hindberg, Hilde},
  title        = {A method for automated snow avalanche debris detection through use of synthetic aperture radar ({SAR}) imaging},
  journaltitle = {Earth and Space Science},
  year         = {2016},
  volume       = {3},
  number       = {11},
  pages        = {446--462},
  doi          = {10.1002/2016EA000168}
}

@article{Eckerstorfer2019,
  author       = {Eckerstorfer, Markus and Vickers, Hannah and Malnes, Eirik and Grahn, Jakob},
  title        = {Near-real time automatic snow avalanche activity monitoring system using {Sentinel-1 SAR} data in Norway},
  journaltitle = {Remote Sensing},
  year         = {2019},
  volume       = {11},
  number       = {23},
  eid          = {2863},
  doi          = {10.3390/rs11232863}
}

@article{Bianchi2021,
  author       = {Bianchi, Filippo Maria and Grahn, Jakob and Eckerstorfer, Markus and Malnes, Eirik and Vickers, Hannah},
  title        = {Snow avalanche segmentation in {SAR} images with fully convolutional neural networks},
  journaltitle = {IEEE Journal of Selected Topics in Applied Earth Observations and Remote Sensing},
  year         = {2021},
  volume       = {14},
  pages        = {75--82},
  doi          = {10.1109/JSTARS.2020.3036914}
}

@article{Bartelt2002,
  author       = {Bartelt, Perry and Lehning, Michael},
  title        = {A physical {SNOWPACK} model for the Swiss avalanche warning: Part I: Numerical model},
  journaltitle = {Cold Regions Science and Technology},
  year         = {2002},
  volume       = {35},
  number       = {3},
  pages        = {123--145},
  doi          = {10.1016/S0165-232X(02)00074-5}
}

@inproceedings{Herla2024,
  author    = {Herla, Florian and Widforss, Aron and Binder, Michael and M{\"u}ller, Karsten and Horton, Simon and Reisecker, Michael and Mitterer, Christoph},
  title     = {Establishing an operational weather \& snowpack model chain in Norway to support avalanche forecasting},
  booktitle = {Proceedings of the International Snow Science Workshop 2024},
  year      = {2024},
  location  = {Troms{\o}, Norway},
  pages     = {168--175},
  url       = {https://arc.lib.montana.edu/snow-science/item/3129},
  urldate   = {2026-08-31}
}

@article{Buser1983,
  author       = {Buser, O.},
  title        = {Avalanche forecast with the method of nearest neighbours: An interactive approach},
  journaltitle = {Cold Regions Science and Technology},
  year         = {1983},
  volume       = {8},
  number       = {2},
  pages        = {155--163},
  doi          = {10.1016/0165-232X(83)90006-X}
}

@inproceedings{Kronholm2006,
  author    = {Kronholm, Kalle and Vikhamar-Schuler, Dagrun and Jaedicke, Christian and Isaksen, Ketil and Sorteberg, Asgeir and Kristensen, Krister},
  title     = {Forecasting snow avalanche days from meteorological data using classification trees: Grasdalen, western Norway},
  booktitle = {Proceedings of the International Snow Science Workshop 2006},
  year      = {2006},
  location  = {Telluride, Colorado, USA},
  pages     = {786--795},
  url       = {https://arc.lib.montana.edu/snow-science/item/1016},
  urldate   = {2026-08-31}
}

@article{Pozdnoukhov2011,
  author       = {Pozdnoukhov, A. and Matasci, G. and Kanevski, M. and Purves, R. S.},
  title        = {Spatio-temporal avalanche forecasting with {Support Vector Machines}},
  journaltitle = {Natural Hazards and Earth System Sciences},
  year         = {2011},
  volume       = {11},
  number       = {2},
  pages        = {367--382},
  doi          = {10.5194/nhess-11-367-2011}
}

@inproceedings{Harvey2016,
  author    = {Harvey, Stephan and van Herwijnen, Alec and Richter, Bettina},
  title     = {Statistical nowcast of avalanche activity at the regional scale},
  booktitle = {Proceedings of the International Snow Science Workshop 2016},
  year      = {2016},
  location  = {Breckenridge, Colorado, USA},
  pages     = {1173--1179},
  url       = {https://arc.lib.montana.edu/snow-science/item/2437},
  urldate   = {2026-08-31}
}

@article{Sielenou2021,
  author       = {Dkengne Sielenou, Pascal and Viallon-Galinier, L{\'e}o and Hagenmuller, Pascal and Naveau, Philippe and Morin, Samuel and Dumont, Marie and Verfaillie, Deborah and Eckert, Nicolas},
  title        = {Combining random forests and class-balancing to discriminate between three classes of avalanche activity in the French Alps},
  journaltitle = {Cold Regions Science and Technology},
  year         = {2021},
  volume       = {187},
  eid          = {103276},
  doi          = {10.1016/j.coldregions.2021.103276}
}

@article{PerezGuillen2022,
  author       = {P{\'e}rez-Guill{\'e}n, Cristina and Techel, Frank and Hendrick, Martin and Volpi, Michele and van Herwijnen, Alec and Olevski, Tasko and Obozinski, Guillaume and P{\'e}rez-Cruz, Fernando and Schweizer, J{\"u}rg},
  title        = {Data-driven automated predictions of the avalanche danger level for dry-snow conditions in Switzerland},
  journaltitle = {Natural Hazards and Earth System Sciences},
  year         = {2022},
  volume       = {22},
  pages        = {2031--2056},
  doi          = {10.5194/nhess-22-2031-2022}
}

@article{Viallon2023,
  author       = {Viallon-Galinier, L{\'e}o and Hagenmuller, Pascal and Eckert, Nicolas},
  title        = {Combining modelled snowpack stability with machine learning to predict avalanche activity},
  journaltitle = {The Cryosphere},
  year         = {2023},
  volume       = {17},
  pages        = {2245--2260},
  doi          = {10.5194/tc-17-2245-2023}
}

@article{Mayer2023,
  author       = {Mayer, Stephanie and Techel, Frank and Schweizer, J{\"u}rg and van Herwijnen, Alec},
  title        = {Prediction of natural dry-snow avalanche activity using physics-based snowpack simulations},
  journaltitle = {Natural Hazards and Earth System Sciences},
  year         = {2023},
  volume       = {23},
  pages        = {3445--3465},
  doi          = {10.5194/nhess-23-3445-2023}
}

@article{Hendrick2023,
  author       = {Hendrick, Martin and Techel, Frank and Volpi, Michele and Olevski, Tasko and P{\'e}rez-Guill{\'e}n, Cristina and van Herwijnen, Alec and Schweizer, J{\"u}rg},
  title        = {Automated prediction of wet-snow avalanche activity in the Swiss Alps},
  journaltitle = {Journal of Glaciology},
  year         = {2023},
  volume       = {69},
  number       = {277},
  pages        = {1365--1378},
  doi          = {10.1017/jog.2023.24}
}

@article{Eiselt2025,
  author       = {Eiselt, Kai-Uwe and Graversen, Rune Grand},
  title        = {Predicting avalanche danger in northern Norway using statistical models},
  journaltitle = {The Cryosphere},
  year         = {2025},
  volume       = {19},
  pages        = {1849--1871},
  doi          = {10.5194/tc-19-1849-2025}
}

\end{document}